\documentclass[10pt,twocolumn,letterpaper]{article}

\usepackage{wacv}
\usepackage{times}
\usepackage{epsfig}
\usepackage{graphicx}
\usepackage{amsmath}
\usepackage{amssymb}
\usepackage{algpseudocode}
\usepackage{algorithm}
\usepackage{multirow}
\DeclareMathOperator*{\argmax}{arg\,max}
\DeclareMathOperator*{\argmin}{arg\,min}
\usepackage[normalem]{ulem}

\def\wacvPaperID{1211} % Enter the WACV Paper ID here

\wacvfinalcopy % *** Uncomment this line for the final submission
\ifwacvfinal
\def\assignedStartPage{1} % *** Enter the assigned starting page number (instead of 9876)
\fi

\ifwacvfinal
\usepackage[breaklinks=true,bookmarks=false]{hyperref}
\else
\usepackage[pagebackref=true,breaklinks=true,colorlinks,bookmarks=false]{hyperref}
\fi

\ifwacvfinal
\else
\fi

\usepackage{authblk}
\begin{document}

%%%%%%%%% TITLE
\title{Active Learning for Bayesian 3D Hand Pose Estimation}
\author[1]{Razvan Caramalau}
\author[1]{Binod Bhattarai}
\author[1,2]{Tae-Kyun Kim}
\affil[1]{Imperial College London, UK }
\affil[2]{KAIST, South Korea}
\affil[ ]{{\tt\small \{r.caramalau18, b.bhattarai, tk.kim\}@imperial.ac.uk}}
% \author[1]{Razvan Caramalau}
% \author[1,2]{Tae-Kyun Kim}
% \affil[1]{\footnotesize Imperial College London}
% \affil[2]{\footnotesize KAIST}
% \author{Razvan Caramalau \quad Binod Bhattarai \quad Tae-Kyun Kim \\
% Imperial College London, KAIST\\
% {\tt\small \{r.caramalau18, b.bhattarai, tk.kim\}@imperial.ac.uk}
% % \and
% % Second Author\\
% % First line of institution2 address\\
% % {\tt\small secondauthor@i2.org}
% }

\maketitle
%\thispagestyle{empty}

%%%%%%%%% ABSTRACT
\begin{abstract}
   We propose a Bayesian approximation to a deep learning architecture for 3D hand pose estimation. Through this framework, we explore and analyse the two types of uncertainties that are influenced either by data or by the learning capability. Furthermore, we draw comparisons against the standard estimator over three popular benchmarks. The first contribution lies in outperforming the baseline while in the second part we address the active learning application. We also show that with a newly proposed acquisition function, our Bayesian 3D hand pose estimator obtains lowest errors with the least amount of data. The underlying code is publicly available at: \url{https://github.com/razvancaramalau/al_bhpe}.
\end{abstract}

%%%%%%%%% BODY TEXT
\section{Introduction}
\label{sec:intro}
Hand Pose Estimation (HPE) is an important research topic where a learning algorithm maps from images the coordinates of the hand skeleton. 
With the recent advancement in deep learning \cite{deepprior,deepprior++,Baek_2018_CVPR,Wan_2018_CVPR,Moon_2018_CVPR,qiye,icvl}, pose estimation has become one of the key ingredients in Robotics, Augmented Reality (AR)/ Virtual Reality (VR), Human-Computer Interaction (HCI) and to mention but a few. In this work, we address the problem of 
Bayesian 3D hand pose estimation in the active learning setting. We consider a scenario where the hand skeleton is represented by volumetric information, captured by depth cameras where there is a limited budget to annotate the hand skeleton. The downstream task is to estimate 3D coordinates of the pre-defined key locations of the hands. 
% representation consisting of the volumetric representation of the hand skeleton from depth-camera images. 
The advantages of depth-based representation being illumination and colour invariant motivated development of multiple large-scale depth-based benchmarks \cite{bighand,icvl,nyu} and several 
international challenges \cite{hands2017,HIM2017,hands2019}.

% It is applied in wide ranges of applications such as AR/VR~\cite{}, Robotic Grasping~\cite{}.
Most of the success stories on 3D-HPE~\cite{hands2017,HIM2017} are due to models with a large number of learnable parameters and with the availability of large-scale annotated data (in the order of $10^6$) such as BigHand2.2~\cite{bighand}.  Annotating such datasets requires a lot of effort, is expensive and also time-consuming. Therefore, it is essential to develop methodologies that identify a small sub-set of the most influencing and discriminative examples to annotate. Active Learning (AL) frameworks~\cite{Sinha2019VariationalLearning,Gal2016} have widely been used for such a purpose.

\begin{figure}
    \centering
    \includegraphics[trim= 0.5cm 0cm 0.5cm 0cm, clip, width=\linewidth]{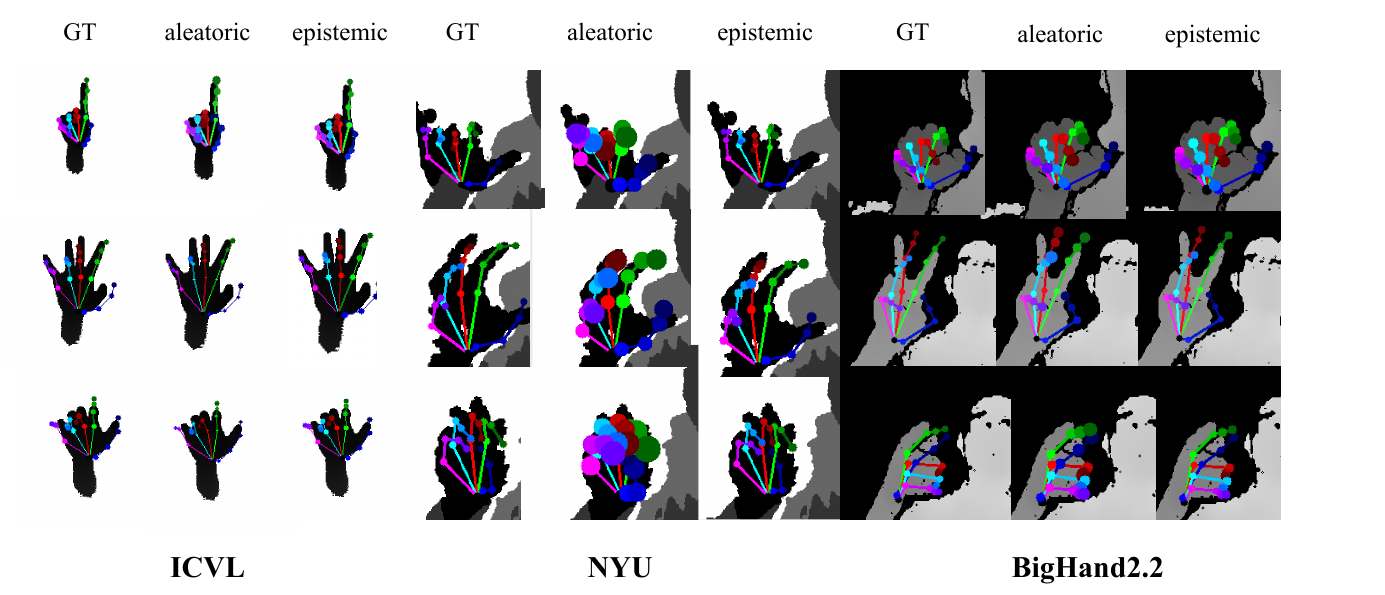}
    \caption{Hand depth images and their corresponding skeleton annotation pairs in ICVL, NYU and BigHand2.2 data sets. (GT- ground truth; aleatoric/epistemic - predicted 3D hand poses with their corresponding uncertainties defined by joint circle radii)}
    \label{fig:intro_fig}
\end{figure}

The AL framework is a well-studied research domain applied for 
several tasks such as image classification~\cite{Gal2016, BeluchBcai2018TheClassification,Pinsler2019BayesianApproximation}, semantic segmentation~\cite{Sinha2019VariationalLearning, viewal}, human pose estimation~\cite{Liu2017a, Yoo2019LearningLearning}. However, it has not yet been applied to the 3D HPE. In this paper, we systematically adapt the classical architecture of DeepPrior\cite{deepprior} in a Bayesian Convolutional Neural Network, Bayesian DeepPrior. Additionally, we present a novel AL selection method, the key component of this framework, optimised for the Bayesian learner. Finally, we evaluate it on three challenging benchmarks for 3D HPE.
% \textcolor{blue}{Active learning has shown to be an efficient and effective strategy in reducing the training set while maintaining the desired model's performance. Thus, it is bridging the gap between the model's learning capability and the representatives of the acquired dataset. In pose estimation, active learning plays a critical role as the annotation stage is often laborious, hardly automated and sometimes imprecise \cite{bighand,icvl,hands2019}. Moreover, in 3D hand pose estimations, datasets might lack viewpoint and articulation variation while storing redundant poses\cite{HIM2017}. 
% }

AL frameworks primarily consist of two major components: learner and sampling technique. 
With the increasing trend of deep learning algorithms usage, the learner is approximated by a large-scale standard CNN, DeepPrior~\cite{deepprior}. However, these frameworks
ignore modelling important uncertainties incurred due to either noisy acquired data (\emph{aleatoric}). 
and due to model's lack of knowledge (\emph{epistemic}). Modelling these uncertainties on discriminative models for
semantic segmentation~\cite{segnet} and depth regression~\cite{uncertainties}, has proven to be effective. As AL 
frameworks are principally designed to select the most influencing and discriminative examples, it is crucial to model both uncertainties. To this end, we propose to approximate the learner by replacing standard DeepPrior with its Bayesian adapted version, similarly to ~\cite{segnet}. To the best of our knowledge, this is the first work to employ a Bayesian 3D-HPE as a learner in an AL framework. The sampling technique is another important component to determine the fate of the AL framework. Existing acquisition functions such as Coreset~\cite{Sener2017ActiveApproach} are widely and successfully used. However, the major limitations of this method consists in relying only on the fixed mean posterior probability while ignoring its epistemic variance. This mean value does not describe fully the complete characteristics of the predicted skeleton. Hence, we propose a novel sampling technique called CKE (a combination between CoreSet and epistemic uncertainty) which models both the upper and lower bound of the epistemic variance of the averaged predicted skeleton. Figure~\ref{fig:intro_fig} shows some of the hand's depth images and their skeleton annotations together with their corresponding uncertainties (represented as circle radii). We describe in more details BraIn our idea in Section~\ref{method}. 

% The training involves estimating the parameters of these 
% components alternatively. To elaborate, the learner is trained to estimate the downstream task, which is 3D HPE for us and the sampler selects the most discriminative examples to query their labels. 

% Our contributions lies on the both the components. Figure~\ref{fig:pipeline} shows the 
% overall end-to-end pipeline of the proposed method. The first
% component Bayesian Deep Network constitutes the leaner of the active learning framework. 
% Unlike some of the recent active learning frameworks~\cite{} where the learner is implemented by
% the Deep CNN, we proposed to approximate it  Bayesian Deep Network~\cite{}. 
% Recent studies on semantic segmentation~\cite{segnet} based on Bayesian Deep Network 
% have shown a lot of promises/advantages over the standard CNN. It has been more important to know 
% what model does not understand or uncertain about~\cite{} rather.... 
% % on several computer vision tasks such as Semantic Segmentation~\cite{}. 

% Things to highlight
% \begin{itemize}
%     \item Bayesian NN as a learner but why we need such backbone? Why do we need to model the uncertainities?  
%     \item first work for active learning in 3D hand pose estimation
%     \item A Novel Sampler and its advantages over the Coreset
% \end{itemize}

To summarise the contributions:

\begin{itemize}
    \item We formulate the 3D Hand Pose estimation problem under the active learning framework.
    \item We propose to approximate in a Bayesian fashion the DeepPrior 3D-HPE.
    \item Proposed a novel AL sampling technique incorporating both the predicted skeleton and its epistemic variance.
    \item We systematically evaluated active learning for both standard and Bayesian DeepPrior on three challenging 3D-HPE benchmarks: BigHand2.2, ICVL and NYU.
    \item The proposed method consistently outperforms the counter-part competitive baselines.
\end{itemize}

\section{Related Work}
\label{related_work}
\begin{figure*}
    \centering
     \includegraphics[trim=0.2cm 0cm 0.01cm 0cm, clip, width=0.8\linewidth]{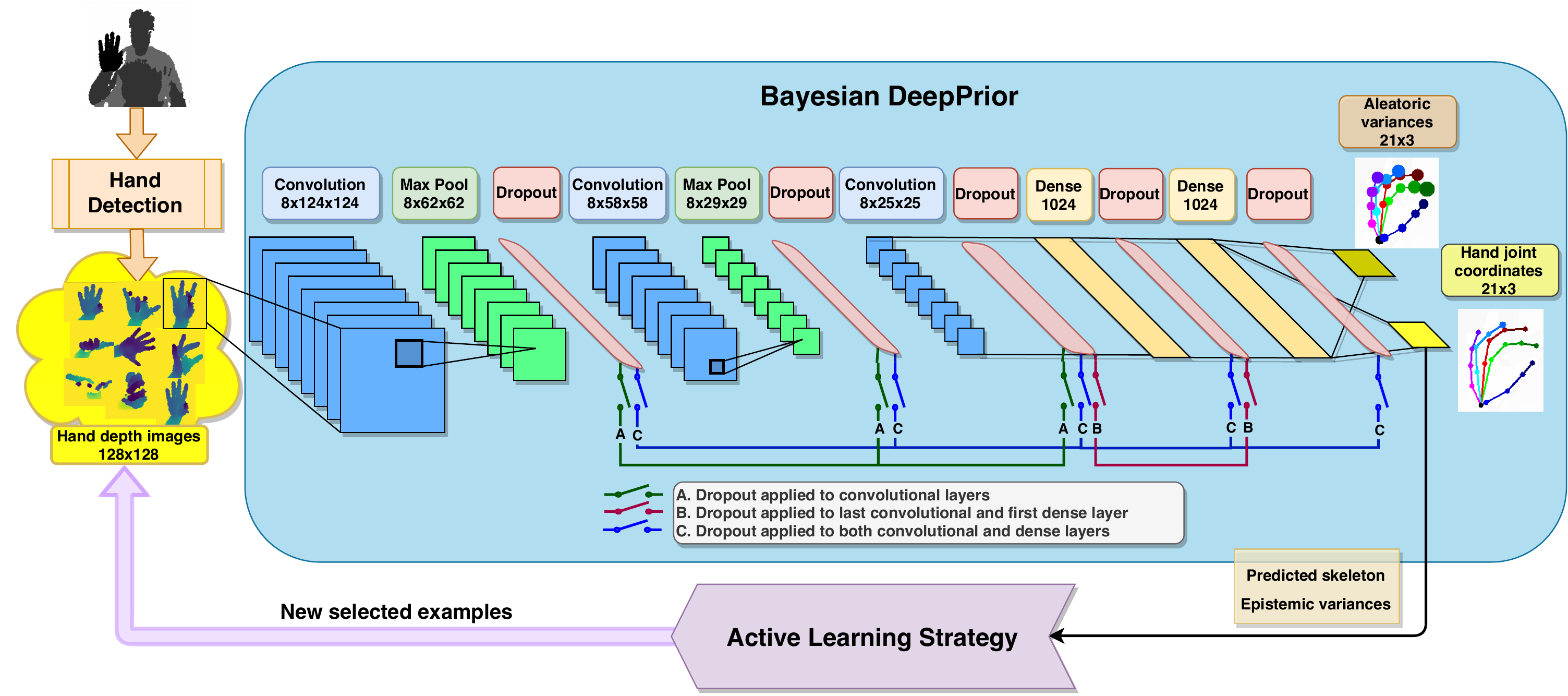}
    \caption{Schematic diagram showing the end-to-end pipeline of the proposed method. The depth image of the hand is pre-processed and fed to the Bayesian DeepPrior Network. This network is trained to minimise objective function given in Equation.\textcolor{red}{2}. The aleatoric and epistemic uncertainties of a sample that we obtain from $M$ number of Monte Carlo Dropouts are passed on to the selection criteria for querying unlabelled examples. The selection method is highlighted in the Algorithm~\ref{alg:cke}. The switches in the Figure are present to indicate the activation of Dropout in different Bayesian approximations.}
    \label{fig:pipeline}
\end{figure*}
\noindent \textbf{3D Hand Pose Estimation.} The first comprehensive review in hand pose estimation that established the current taxonomy is published in \cite{EROL200752}. Together with the deep learning popularity and easy access to depth camera sensors, 3D HPE has gained a deep interest in the computer vision community. In terms of methodology, theoretical approaches have investigated 2D or 3D representations, detection-based~\cite{Moon_2018_CVPR}, hierarchical and/or structured models~\cite{qiye,deepprior,icvl} over single or multi-stage processes. Earlier state-of-the-art methods favoured combination of discriminative (random forest \cite{icvl} or CNN-based\cite{deepprior}) and generative solutions \cite{Baek_2018_CVPR} as in \cite{qiye,deepprior++}. Due to the past concepts from \cite{pointnet}, volumetric representations \cite{Moon_2018_CVPR} managed to out-stand by compensating from the high non-linearity in direct regressions. These recent works \cite{Moon_2018_CVPR,Wan_2018_CVPR,a2j} have obtained impressive results with average 3D joint errors below 10 mm on datasets like NYU~\cite{nyu}, ICVL~\cite{icvl} or BigHand2.2~\cite{bighand}. As we are tackling uncertainty exploration and data representatives of the model, for simplicity and efficient analysis we deploy a standard DeepPrior\cite{deepprior} architecture. Even though the accuracy is lower than current arts due to relatively lesser model parameters, our method is generic, the insights obtained can be easily transferred to deeper and generative models.\\
\noindent \textbf{Bayesian Deep Learning.} Recently, there have been several investigations in quantifying and representing uncertainty in CNNs. A novel approach is to approximate variational inference in a Bayesian implementation \cite{gal2015bayesian} for image classification\cite{Gal2016}. Furthermore, the types of uncertainties (aleatoric or epistemic) and the ways to employ these uncertainties in both regression and classification tasks are presented in \cite{uncertainties, unc2018}. The novel  approximation of Bayesian\cite{Gal2016DropoutGhahramani} for deep learning by using the Dropout layers\cite{dropout} reduced the computational complexities of the na\"{i}ve Bayesian Neural Network implementation. An analysis of uncertainty for the active learning framework is presented in ~\cite{zeng2018relevance}. However, the evaluation is conducted on small scale dataset such as MNIST \cite{mnist} classification . In contrast to this, our work analyses a more difficult and  large-scale experimental setup.
% It is not necessarily transferable to other large-scale vision tasks. 
Another recent work on image classification~\cite{BeluchBcai2018TheClassification} proposed ensemble-based active learning and demonstrates outperforming the Dropout Bayesian approximation\cite{Gal2016DropoutGhahramani}. Again the experiments are constrained on small scale scenarios. 
On these premises, we further analyse the exploration of both data and model-dependent uncertainties similar to \cite{uncertainties}, but for the large-scale and more challenging problem i.e. 3D-HPE. Furthermore, this will integrate the risk measurement concerns presented in \cite{Osband2016RiskVU} through the aleatoric uncertainty.\\
\noindent \textbf{Active Learning Methods.} This branch of machine learning was explored in the need of informative datasets that are in most cases model-dependent. A survey of different standard active learning schemes is in~\cite{settles.tr09}. With the advances in deep neural networks, the research community began to integrate the classical approaches despite the lack of integration in online training (essential part of the active learner methodology). The most common scenario used, pool-based sampling, limits the live model refinement by performing offline training\cite{Sener2017ActiveApproach, Gal2017DeepDatab, Yoo2019LearningLearning,Sinha2019VariationalLearning}.
On the other hand, pool-based active learning opened a new direction of research for deep neural networks~\cite{Gal2016}. This has evaluated with what minimum percentage of the training set the model can achieve the same accuracy as the entire one. 
In terms of pose estimation, a practical application of active learning for hands has been conducted in \cite{proactiveses}, where KD-trees are applied to guide the camera movement to a more informative viewpoint. A theoretical approach has been explored in~\cite{Liu2017a} over the human pose estimation problem by actively collecting unlabelled data from the heat-map output of the Convolutional Pose Machines (CPMs). However, our methodology is driven independently from the output as it relies directly on the predicted hand skeleton.
% Add our methodology where it stands

\section{Method}
\label{method}

% Description of the pipeline and analysis
% In this section, we firstly empirically construct the Bayesian approximation of a standard 3D HPE baseline similarly to \cite{segnet}. However, we extend the adaptation by exploring the two types of uncertainties defined in Gal.Y. et. al.\cite{uncertainties}. These statistics are derived by inheriting the noisy or poorly annotated data and by observing the learning capability of the model. In the second part, given the Bayesian hand pose architecture, we further investigate acquisition methods in the active learning scheme with the uncertainty variances. Therefore, our contribution relies in the beginning on the analysis of the Bayesian 3D HPE approximation and, finally, in proposing a new active selection mechanism that outperforms other state-of-the-art techniques.
We start this section with formulating the 3D-HPE problem as a Bayesian approximation inspired by ~\cite{segnet,uncertainties} for semantic segmentation and depth regression. We adopt such a framework (similarly to Kendall et. al. in \cite{uncertainties}) to model the two uncertainties: aleatoric and epistemic. These two statistics characterise the important, but complementary aspect of the model when trained from data. In particular, aleatoric uncertainty captures uncertainty due to noisy training examples which can not be eradicated from the model even if there is plenty of the training examples. Whilst, epistemic uncertainty quantifies the ignorant aspect of the model parameters which can be addressed with the availability of training examples. For an AL application, the aleatoric uncertainty plays a key informative role, to avoid annotating difficult or noisy samples when acquiring new data. Besides, the epistemic variance helps the AL sampling method in indicating relevant unseen data for the learner.
In the second part, given the parameters inferred from the Bayesian hand pose estimator, we further investigate an acquisition method,  commonly known as a sampler in the active learning scheme, by enclosing the uncertainty variances. Hence, our
contribution relies on the analysis of the Bayesian 3D-HPE approximation (comprises the learner component in AL), together with a newly proposed selection mechanism. Figure~\ref{fig:pipeline} summarises the proposed pipeline.
% that outperforms other state-of-the-art techniques.

\subsection{Bayesian 3D Hand Pose Estimator}
\subsubsection{3D Hand Pose Estimator}
In classification tasks, uncertainty is estimated by the posterior probability of a class. However, the 3D hand pose estimation, a regression problem, maps hand (depth) images to 3D coordinates of the hand joint locations. In our scenario,
hand depth-image and representative coordinates to describe its skeleton are given. We describe the hand skeleton following~\cite{EROL200752} in 21 joints.
With this information, we shortly describe a method to regress the skeleton coordinates along with modelling the mentioned uncertainties.
 
% \sout{deals mapping to 3D coordinates of the hand joint locations from images captured with depth cameras. The standard pipeline of this task \sout{assumes} \textcolor{blue}{consists of} a hand detection algorithm as a front-end to the estimator. Therefore, the inputs to the 3D HPE are cropped images centered on the hand region. Reducing the potential cluttered or noisy areas in the depth scene motivates the main advantage of this practice.
% We describe the hand skeletons following~\cite{EROL200752} by 21 volumetric joints: 1 root palm, 5 metacarpophalangeals (MCP), 5 proximal interphalangeals (PIP), 5 distal interphalangeals (DIP)  and 5 finger tips.}

\def \R {\mathbb{R}}
\def \x {\mathbf{x}}
\def \y {\mathbf{y}}
\def \ssigma {\mathbf{\sigma}}
In order to topologically evaluate the regression uncertainties, we deploy a standard \emph{DeepPrior} \cite{deepprior} architecture. This comprises of a convolutional feature extractor and a dense regression. Given $\x \in \R^{w \times h}$, a 2D cropped hand image with $w$ width and $h$ height is inferred through the CNN, together with its corresponding ground-truth $\y \in \R^{21 \times 3}$. Structurally, the feature extractor is composed of three block groups of convolution, max pool and LeakyReLU activation. The flattened output of the feature extractor is regressed in the end by 2 dense layers (see Figure \ref{fig:pipeline}).
For simplicity, we exclude the final PCA layer from the initial design of \cite{deepprior} and directly predict the normalised UVD joint coordinates $\hat{\y}$ (description of the pipeline in Fig. \ref{fig:pipeline}). Finally, we optimise the parameters by Stochastic Gradient Descent (SGD) to minimise the mean squared error. The objective function is as below:
\begin{equation}
    \label{eg1}
    \mathcal{L}(\x, \y; \Theta) = \frac{1}{n}\sum_{i=1}^n\Big( \frac{1}{K}\sum_{k=1}^K\|\y_{i,k} - \hat{\y}_{i,k}\|^2\Big),
\end{equation}
where, $n$ is the data size, $\Theta$ represent model parameters and $K$ is the number of joints.

\subsubsection{Bayesian DeepPrior}
\label{bayesian}
Here, we describe the \emph{Bayesian DeepPrior} into details.
As we stated before, unlike in classification model, it is not straight forward to model the uncertainties in regression model. In order to make the DeepPrior a probabilistic model, we introduced Dropout~\cite{dropout} on its layers similar to ~\cite{Gal2016DropoutGhahramani}.
As in ~\cite{uncertainties}, we also propose to estimate and investigate aleatoric and epistemic uncertainties for our task.
% \sout{Inspired from \cite{Gal2016DropoutGhahramani} to construct the probabilistic model, we introduce Dropout \cite{dropout} layers in the just above mentioned \emph{DeepPrior} architecture as a Bayesian approximation. 
% With the extension of ~\cite{uncertainties} to regression tasks, similarly we generate and investigate two types of uncertainty: a data-dependent, i.e. aleatoric and a model-dependent, i.e. epistemic.}

Bayesian Neural Networks (BNNs) set a Normal distribution $\Theta \sim \mathcal{N}(0,I)$ of their weight parameters as prior. To approximate the posterior distribution over the weights $f(\Theta|\x, \y)$, we minimise the Kullback-Leibler (KL) divergence of a variational inference distribution $q(\Theta)$ and its posterior: KL$(q(\Theta)\| f(\Theta| \x,\y))$. To minimise the KL divergence loss, we apply Monte Carlo Dropout (MCD) during the variational inference and we minimise the mean squared error of joint location's prediction. This minimisation is equivalent to the KL divergence minimisation. For more details, we suggest readers to refer to ~\cite{uncertainties}.
Once we keep Dropout active over the entire $\Theta$ we can obtain a Bayesian approximation of the posterior's  mean and variance.

As  stated in \cite{Osband2016RiskVU}, the uncertainty variance of the BNN consists of the Bayes risk rather than a systematic uncertainty. Therefore, not only do we learn the outputs of the final dense layer, but the aleatoric uncertainty that balances the mean squared error objective function during training as well. Specifically, we allocate an aleatoric variance for each hand joint coordinate. Hence, the new objective function to model such uncertainties is defined as in \cite{uncertainties}:
% \small
\begin{equation}\label{eg:2}
    \begin{split}
    \hspace{-15pt} \mathcal{L}_{B}(\x, \y; \Theta) = \frac{1}{n}\sum_{i=1}^n\Big( \frac{1}{K}\sum_{k=1}^K \frac{1}{2} e^{-\mathbf{\alpha}_{i,k}} \|\y_{i,k} - \hat{\y}_{i,k}\|^2 + \\ 
    & \hspace{-60pt}+ \frac{1}{2} \mathbf{\alpha}_{i,k}\Big),
    \end{split}
\end{equation}
% \normalsize
with the logarithmic variance $\mathbf{\alpha}_{i,k} = \log \hat{\sigma}_{al}^2$ and $\hat{\ssigma}_{al}^2$, the aleatoric variance. Thus, the learnt numerically-stable logarithmic tracks the noise present in the data.

To summarise, by applying Monte Carlo Dropout (MCD) we obtain a Bayesian DeepPrior that generates a mean value for each joint coordinates. After variational inferences we also evaluate its epistemic and learnt aleatoric variances. For $M$ times passes of a sample, the epistemic uncertainty is estimated as below:
\begin{equation}
    \label{eg3}
\hat{\ssigma}_{ep}^2 \approx \frac{1}{M}\sum_{m=1}^M{\mathbf{\hat{y}}_m}^2 - \Big(\frac{1}{M}\sum_{m=1}^M{\mathbf{\hat{y}}_m}\Big)^2
\end{equation}
Finally, the combined variances for a predicted skeleton $\hat{\y}$ can be expressed as:
\begin{equation}
    \label{eg4}
    \hat{\ssigma}(\hat{\y}) \approx \frac{1}{M}\sum_{m=1}^M{\hat{\y}_m}^2 - \Big(\frac{1}{M}\sum_{m=1}^M{\hat{\y}_m}\Big)^2 + \frac{1}{M}\sum_{m=1}^M{\hat{\ssigma}^2_{al_m}}.
\end{equation}
In our experiments section \ref{sec:arch}, we present the study on number of $M$ vs the mean joint error stabilisation.

From an architectural perspective, the Bayesian DeepPrior  contains Dropout layers after every convolutional and dense layer. In practice \cite{segnet,uncertainties}, it has been shown that this model may suffer from strong regularisation. Thus, similarly to \cite{segnet}, we propose different variants where Dropout is applied to designated locations. These probabilistic variants of Bayesian DeepPrior are simulated through switches in the proposed pipeline\ref{fig:pipeline} accordingly: 
\begin{itemize}
\item \textbf{A} - only to all convolutional layers;
\item \textbf{B} - centrally to the last convolutional layer and the first dense layer;
\item \textbf{C} - throughout the entire DeepPrior architecture.
\end{itemize}
To fine-tune the design of the Bayesian DeepPrior architecture, we perform cross-validation study on the NYU Hand dataset \cite{nyu} in section \ref{sec:arch}. 

\def \s {\mathbf{s}}
\subsection{Active Learning Framework}
In this section, we briefly describe the active learning process for deep learning together with the proposed acquisition function adapted for the Bayesian DeepPrior. \\
\noindent \textbf{Pool-based Active Learning Strategy.}
As active learning has gained stronger interest in deep learning, a pool-based scenario has become a standard methodology to overcome the data-greedy models with their slow training process~\cite{settles.tr09}. Therefore, the pool-based active learning considers a scenario with an initial annotated set $\s^0$ and an available unlabelled dataset $U_{pool}$. The goal is to find the least amount of $L$ annotated subsets $\s^1,\s^2 \dots \s^L \subset U_{pool}$ so that we achieve the targeted mean squared joint error. Given an acquisition function $\mathcal{A}$, this can be summarised under the following equation:
\begin{equation}
    \label{eg5}
    \min_L \min_{\mathcal{L}_B}\mathcal{A}(\mathcal{L}_B; \s^1,\s^2 \dots \s^L \subset U_{pool}).
\end{equation}
In the next section, we analyse and propose a function $\mathcal{A}$ suitable for our Bayesian DeepPrior architecture.\\
\begin{figure}
    \centering
     \includegraphics[trim=0cm 0cm 0cm 2cm, clip, width=1.0\linewidth]{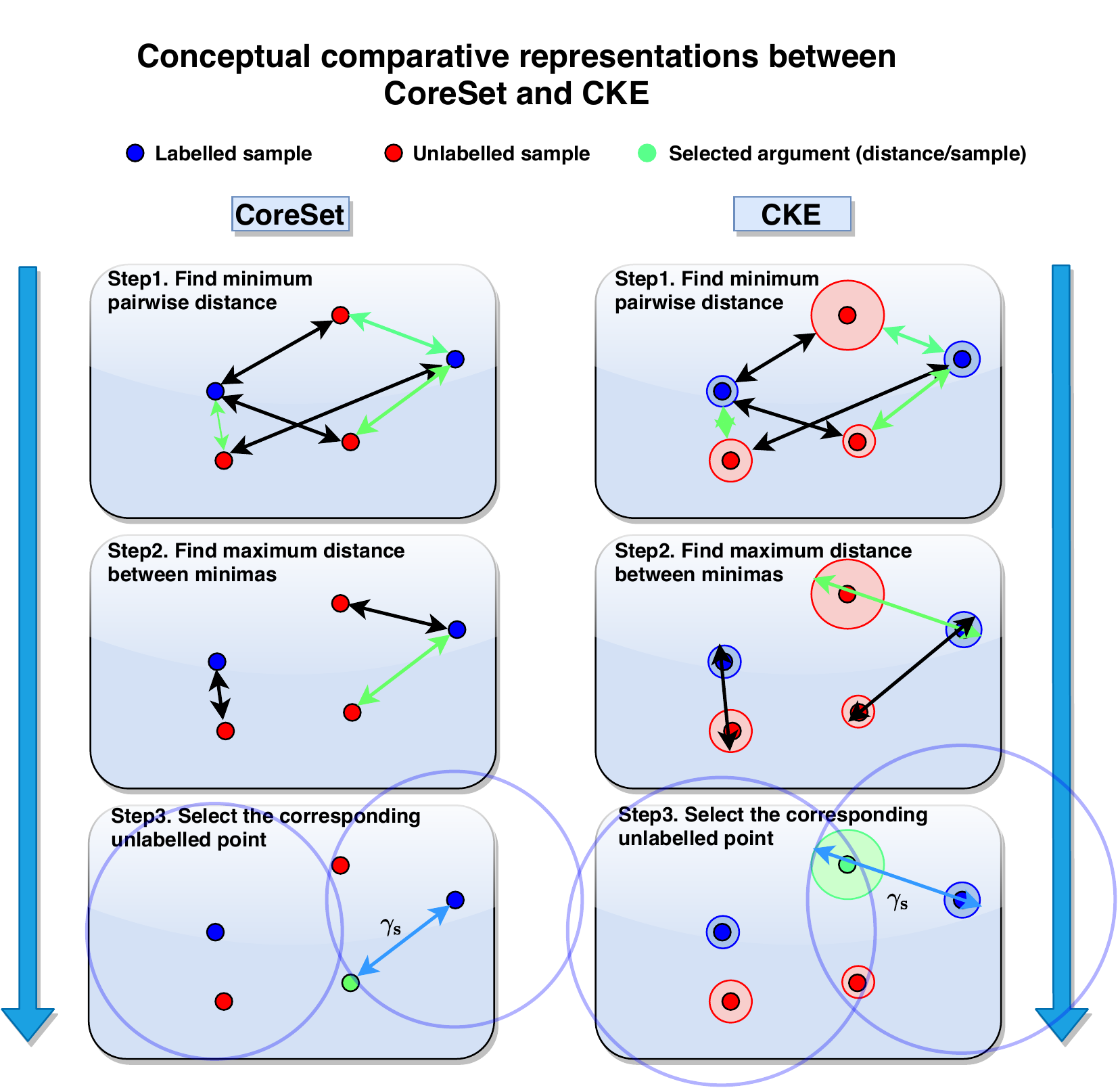}
    \caption{Conceptual comparison between the sampling techniques: Coreset\cite{Sener2017ActiveApproach} vs CKE. The blue dots represent the annotated samples in the hand skeleton space while red are to be selected. The circle radii in the CKE process stages are defined by each samples' uniform distribution. }
    \label{fig:cke}
\end{figure}
% \subsubsection{Combined geometric and uncertainty acquisition function}
\noindent \textbf{Combination of Geometric and Uncertainty Acquisition Function.}
\label{sec:cke}
Acquisition functions have been extensively developed for classification tasks \cite{Gal2017DeepDatab,Pinsler2019BayesianApproximation} due to their probabilistic output. However, in regressions, we either have to derive statistics like in \cite{Liu2017a,uncertainties} or develop separate data analysis through task-invariant methods \cite{Sinha2019VariationalLearning,Yoo2019LearningLearning,Sener2017ActiveApproach}.
The benefit of our Bayesian approximation is that we can use the epistemic and/or aleatoric uncertainties to filter out the unlabelled pool of data. Apart from the uncertainty variances, we also propose to revise the \emph{CoreSet} \cite{Sener2017ActiveApproach} acquisition function under our Bayesian methodology.

In principle, CoreSet treats the minimisation of the geometric bound between the objective functions of the annotated set $\s^0$ and of a representative subset $\s$ from the unlabelled examples. The bounding between the two losses applied in our context can be expressed as:
\begin{equation}
    \label{eg6}
\Big| \mathcal{L}_B(\mathbf{x}_i,\mathbf{y}_i \in \s)-\mathcal{L}_B(\mathbf{x}_j,\mathbf{y}_j \in \s^0)\Big|\leq \mathcal{O}(\gamma_\mathbf{s})+\mathcal{O}\Big(\sqrt{\frac{1}{B}}\Big),
\end{equation}
where $\gamma_\mathbf{s}$ is the fixed cover radius over the entire data space and $B$ is the budget of samples to annotate.
It has been shown in \cite{Sener2017ActiveApproach} that this risk minimisation can be approximated through the  k-Centre Greedy optimisation problem \cite{wolf}. As it relies on $\Delta$, the $l2$ distances between samples of posterior distribution, we define the selection under the following scope:
\begin{equation}
    \label{eq7}
    \argmax_{i\in\s}\min_{j \in \s^0}\Delta(\hat{\y}_i,\hat{\y}_j).
\end{equation}

Considering the Bayesian DeepPrior architecture, we extend the CoreSet solution by including the epistemic variance in the distance computation from equation \ref{eq7}. Moreover, the estimated 3D coordinates are averaged after $M$ MCD inferences. Therefore, when evaluating $\min_{j \in \s^0}\Delta(\hat{\y}_i,\hat{\y}_j)$, we subtract their corresponding standard deviations $\hat{\ssigma}_{ep_i}$ and $\hat{\ssigma}_{ep_j}$ so that closer uncertain centres are highlighted. Implicitly, we extend with $\hat{\ssigma}_{ep_i}$ and $\hat{\ssigma}_{ep_j}$ these minimum pairwise distances ($\min_{j \in \s^0}\Delta(\hat{\y}_i,\hat{\y}_j)$) when we compute the maximum distance from the unlabelled. Finally, we add to the subset $\s^0$ the furthest uncertain sample to be annotated. The impact of the epistemic uncertainty variance is adjusted with a parameter $\eta$. We repeat this number of steps according to a budget. 

We define this adapted combination  between the k-Centre Greedy algorithm and the epistemic variances as \textbf{CKE}. The pseudo-code from Algorithm \ref{cke} presents the steps for selecting a subset $B_\mathbf{ublb}$, given a budget $B$. Furthermore, we conceptually represent in Figure~\ref{fig:cke} the stages of data selections for both CoreSet and CKE. 
\begin{algorithm}
   \caption{CKE}
   \label{cke}
   \begin{algorithmic}[1]
      \State \textbf{Input}: labelled set $\mathbf{x}_j \in \mathbf{s}^0$, unlabeled pool $\mathbf{x}_i \in U_{pool}$, query budget $B$, corresponding epistemic variances $\hat{\sigma}_{ep_i}$ and  $\hat{\sigma}_{ep_j}$
      \State Initialise $\s \subset U_{pool}$
      
      \Repeat
       \State $\Delta \mathbf{d_{ub}}_{i} = \min_{j \in \s^0} \Delta(\mathbf{\hat{y}}_i+\frac{\eta}{2}\mathbf{\hat{\sigma}}_{ep_i},\mathbf{\hat{y}}_j+\frac{\eta}{2}\mathbf{\hat{\sigma}}_{ep_j})$
      
      \State $\mathbf{arg_{lb}}_{i} = \argmin_{j \in \s^0} \Delta(\mathbf{\hat{y}}_i-\frac{\eta}{2}\mathbf{\hat{\sigma}}_{ep_i},\mathbf{\hat{y}}_j-\frac{\eta}{2}\mathbf{\hat{\sigma}}_{ep_j})$
      
      \State $b = \argmax_{i\in\s} \Delta \mathbf{d_{ub}}_{i} (\mathbf{arg_{lb}}_{i}) $
      
      \State $\mathbf{s_{ublb}} = \mathbf{s}^0 \cup \{b\}$
      
      \Until { $\mathbf{s_{ublb}} = B + \mathbf{s}^0$}
      
      \State \textbf{Return}:  $B_\mathbf{ublb} =  \mathbf{s_{ublb}} \setminus \mathbf{s}^0$
\end{algorithmic}
\label{alg:cke}
\end{algorithm}
We can denote that CKE  benefits of the Bayesian model uncertainty $\hat{\ssigma}^2_{ep}$ when minimising the global geometric cover $\gamma_\s$. In this manner, our proposed solution identifies hand poses furthest from the labelled centres and with the highest epistemic uncertainty deviation. Moreover, the new objective function with learnt aleatoric uncertainties (see Equation \ref{eg:2}) drops the noisy samples making our estimations more robust. On these premises, we consider that our AL method for Bayesian DeepPrior is superior to the standard approach.

\section{Experiments}
\label{experiments}
% \begin{figure*}
%     \centering
%     \includegraphics[trim=1cm 0.5cm 3.5cm 2.5cm, clip, width=0.33\textwidth]{figures/icvl.pdf}
%     \includegraphics[trim=1cm 0.5cm 3.5cm 2.5cm, clip, width=0.33\textwidth]{figures/mega.pdf}
%     \includegraphics[trim=1cm 0.5cm 3.5cm 2.5cm, clip, width=0.33\textwidth]{figures/nyu.pdf}
%     \caption{Empirical comparison I: Comparison of the proposed method with the exisitng active learning methods on ICVL (Left), BigHand (Middle) and NYU (right) data set}
%     \label{fig:al_methods}
% \end{figure*}
\subsection{3D Hand Pose Estimation Datasets} 
% To perform our Bayesian DeepPrior architectural choice together with CKE, the proposed active learning for 3D hand pose estimation, we firstly conduct a brief discussion of the hand datasets. Although the annotation process can be difficult to automatically deploy under noisy environments, there has been a detailed cumulative expansion of the hand data acquisition process. Therefore, we are tackling in the experiment section three well-known hand datasets: ICVL \cite{icvl}, NYU \cite{nyu} and BigHand2.2 \cite{bighand}.
\noindent \textbf{ICVL \cite{icvl}:} This is one of the earliest depth-based hand datasets. It consists of a total number of 17,604 (16,008 train, 1596 test) from 10 subjects with 16 joints annotations. 
% Thus, it is divided in between 16,008 training and 1596 testing images. While there are 10 different subjects used in the acquisition process, ICVL \cite{icvl} still lacks of a high articulation degree under a single frontal viewpoint. However, the annotation process has been done with 16 hand joints tracking and manual tuning by keeping a relative low labelling noise.

\noindent \textbf{NYU\cite{nyu}}. Created from 2 subjects, NYU has 72,757 training images and an 8,252 testing set. The hand skeleton consists of 36 3D coordinates.
% It also presents various hand articulation and rotations at a 640x480 frame size. The hand skeleton is mapped on 36 joint locations. In our analysis, we train and test only the frontal viewpoint as in ICVL.
% This benchmark has been created in a similar fashion to ICVL, but only 2 subjects were considered. Despite that, the dataset is formed of 72,757 training images and a 8,252 testing set. It also presents various hand articulation and rotations at a 640x480 frame size. The hand skeleton is mapped on 36 joint locations. In our analysis, we train and test only the frontal viewpoint as in ICVL.

\noindent \textbf{BigHand2.2\cite{bighand}}. The largest benchmark to date, it consists of 2.2M frames from 10 different hand-models. For practicality, we decide to uniformly sub-sample every 10 frames due to the insignificant accuracy gain over the entire set as analysed in \cite{bighand}. Hence, we train our model with 251,796 frames and test 21 hand key-points on 39,099 frames.  

For all the datasets, we standardise the number of joints to 21 3D locations. We use pre-trained UNet~\cite{unet} to detect hand and centre cropped it to the dimension of 128$\times$128.
% As our contribution lies on the 3D HPE, we pre-process the data by detecting and cropping the hand regions with a pre-trained UNet \cite{unet} architecture and cropped to the dimension of 128$\times$128.
% This will uniformly arrange the 128x128 2D depth inputs for the DeepPrior task.

\subsection{Bayesian DeepPrior Architecture}
% \subsubsection{3D Hand Pose Estimation datasets}
% \noindent \textbf{Bayesian DeepPrior architectural analysis}
\noindent \textbf{Ablation Studies on Architecture.}
\label{sec:arch}
To approximate a Bayesian Neural Network using MCD as described in Section \ref{bayesian}, we propose three probabilistic variants by dropping out different combinations of layers. We applied these three strategies:~\textbf{A} - convolutional feature extractor;~\textbf{B} - centrally, on the last convolution and first dense layer;~\textbf{C} - throughout the entire DeepPrior and evaluated the performance on NYU data set. The comparison of the performance on these configurations is summarised in Table~\ref{Tab:ablation}. Accordingly, when applying MCD only on the feature extractor (\textbf{A}), it yields the best performance. This is because in \textbf{C} Dropout brings too much regularisation, while \textbf{B} locks low-level features in the first convolutional layers.
The trend we observe is similar to one reported in the Bayesian SegNet \cite{segnet}.
We deploy the Adam\cite{Kingma2015ADAM:Optimization} optimiser with a learning rate of $10^{-3}$, a batch size of 128 and a total number of $M=70$ MCDs.  These parameters remain constant for all three setups. 
%create table.
\begin{table}[]
\begin{tabular}{l|l|l|l}
%\hline
Bayesian DeepPrior variants & A                & B       & C       \\ \hline
NYU Testing  MSE {[}mm{]}   & \textbf{22.48} & 22.94 & 22.65 %\\ \hline
\end{tabular}
\caption{Ablation evaluation of Bayesian DeepPrior}
\label{Tab:ablation}
\end{table}
Throughout all of our upcoming experiments, we maintain the Bayesian DeepPrior variant where Dropout is present after the convolutional layers.

% evaluate the ideal number of dropouts.
After identifying the optimal configuration of Dropouts, we played with the number of variational inferences $M$ under the same hyper-parameters setting. We observed that increasing the value of $M$ does not impact the performance, however, adds computational complexity.  Lowering the value of $M$ to 40 does not impact the performance. Hence, we keep this value for the rest of the experiments.  

\noindent \textbf{Bayesian DeepPrior vs Standard DeepPrior }
To demonstrate the effectiveness of our Bayesian DeepPrior approach, we conduct a quantitative comparison against the standard DeepPrior baseline proposed in \cite{deepprior}. We evaluate the average MSE for both train and test sets of the three compared benchmarks. The performance comparison is summarised in Table~\ref{Tab:2}.

\begin{table}[]
\begin{tabular}{c|c|c|c}
\multicolumn{2}{c|}{Hand Dataset}   & DeepPrior & Bayesian DeepPrior \\ \hline
\multirow{2}{*}{ICVL}       & train & 7.1633      & 7.5261              \\ \cline{2-4} 
                            & test  & 10.6233     & \textbf{10.0988}             \\ \hline
\multirow{2}{*}{NYU}        & train & 4.7034    & 7.7566            \\ \cline{2-4} 
                            & test  & 25.0754    & \textbf{22.4838}            \\ \hline
\multirow{2}{*}{BigHand2.2} & train & 12.7731     & 7.7566             \\ \cline{2-4} 
                            & test  & 22.2353   & \textbf{21.4649}   
% 3D HPE       & \multicolumn{2}{c|}{DeepPrior} & \multicolumn{2}{c|}{Bayesian DeepPrior} \\
% Hand Dataset & train          & test          & train              & test               \\ \hline
% ICVL         & 7.1633         & 10.6233       & 7.5261             & \textbf{10.0988}            \\ \hline
% NYU          & 4.7034         & 25.0754       & 7.7566             & \textbf{22.4838}            \\ \hline
% BigHand2.2   & 12.7731        & 22.2353       & 13.8621            & \textbf{21.4649}             
\end{tabular}
\caption{Bayesian DeepPrior vs DeepPrior - averaged MSE [mm]}
\label{Tab:2}
\end{table}

The Bayesian DeepPrior brings a clear advantage over Standard by yielding the lower testing error on all the three benchmarks. 
This shows how effectively we can generalise overall testing sets, while the standard DeepPrior over-fits the training sets. 
In addition to the performance improvement, the Bayesian DeepPrior also generates uncertainty metric for the hand poses.\\
\noindent \textbf{Epistemic and Aleatoric Uncertainties }
\label{ep_al}
% We presented in Section \ref{bayesian} the two type of variances that our proposed model generates. The 3D HPE learns the normalised UVD coordinates of the hand joints. Thus, the epistemic and aleatoric variances have values between 0 and 1. 
For every coordinate of the hand skeleton, our model estimates their aleatoric and epistemic deviation. Table~\ref{Tab:3} enlists the average of both uncertainties for all the three benchmarks. The values are in the range of 0 to 1. Please note, we predict the normalised UVD coordinates of the hand joints.

% We presented in Section \ref{bayesian} the two type of variances that our proposed model generates. The 3D HPE learns the normalised UVD coordinates of the hand joints. Thus, the epistemic and aleatoric variances have values between 0 and 1. Moreover, for each coordinate of the hand skeleton we learn and estimate a corresponding aleatoric and epistemic deviation. In Table 3., we enlist the average of both uncertainties over the three hand benchmarks.

\begin{table}[]
\begin{tabular}{c|c|c|c}
\multicolumn{2}{c|}{Hand Dataset}   & aleatoric var & epistemic var \\ \hline
\multirow{2}{*}{ICVL}       & train & 1.057                                 & 0.004                                \\ \cline{2-4} 
                            & test  & 1.016                                 & 0.0039                               \\ \hline
\multirow{2}{*}{NYU}        & train & 1.169                                 & 0.0142                               \\ \cline{2-4} 
                            & test  & 1.512                                 & 0.0066                               \\ \hline
\multirow{2}{*}{BigHand2.2} & train & 1.964                                 & 0.0132                               \\ \cline{2-4} 
                            & test  & 2.1594                                & 0.0174                              
\end{tabular}
\caption{Averaged epistemic and aleatoric variances at $\times10^{-2}$ on ICVL, NYU and BigHand2.2}
\label{Tab:3}
\end{table}
\begin{figure*}
    \centering
    \includegraphics[trim=1cm 0.5cm 3.5cm 2.5cm, clip, width=0.33\textwidth]{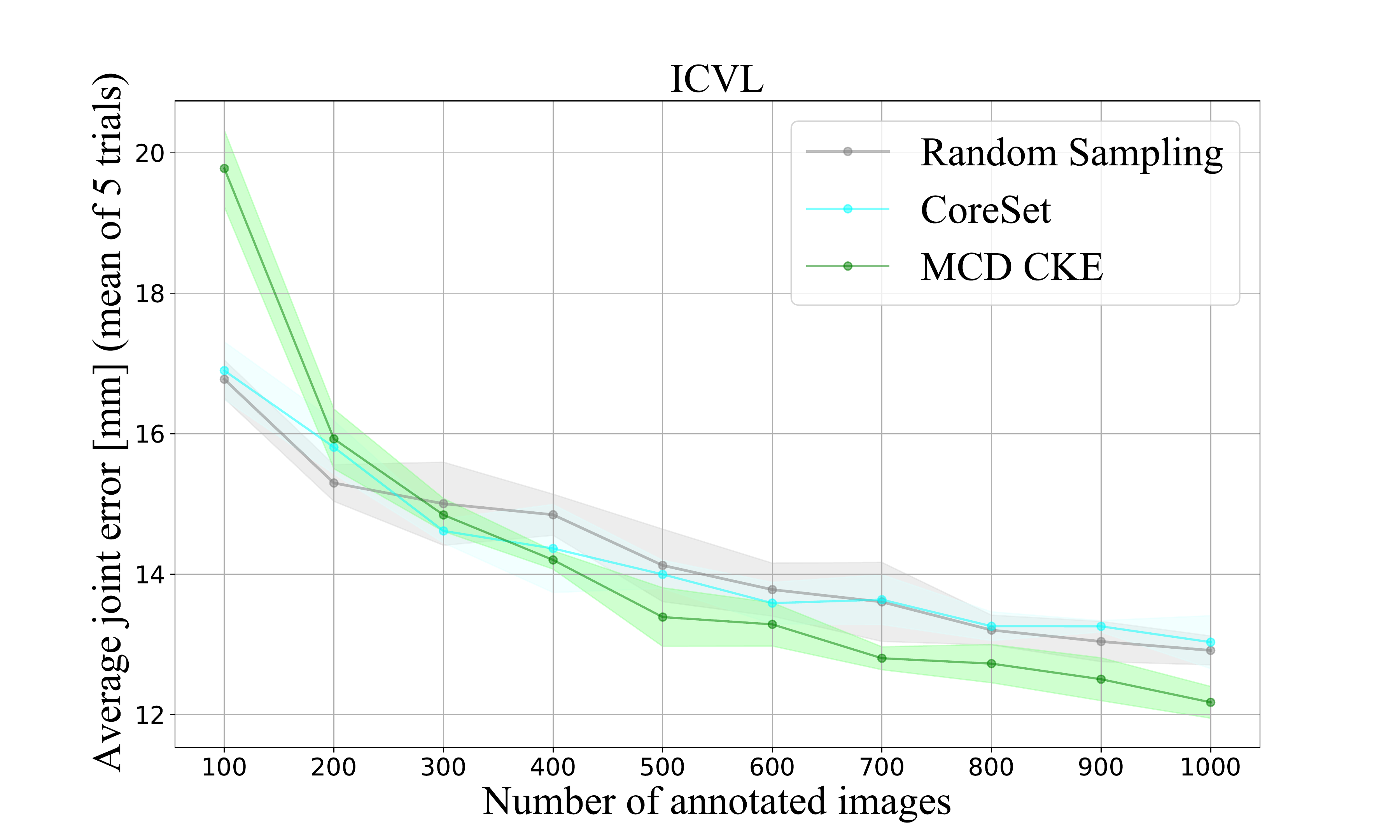}
    \includegraphics[trim=1cm 0.5cm 3.5cm 2.5cm, clip, width=0.33\textwidth]{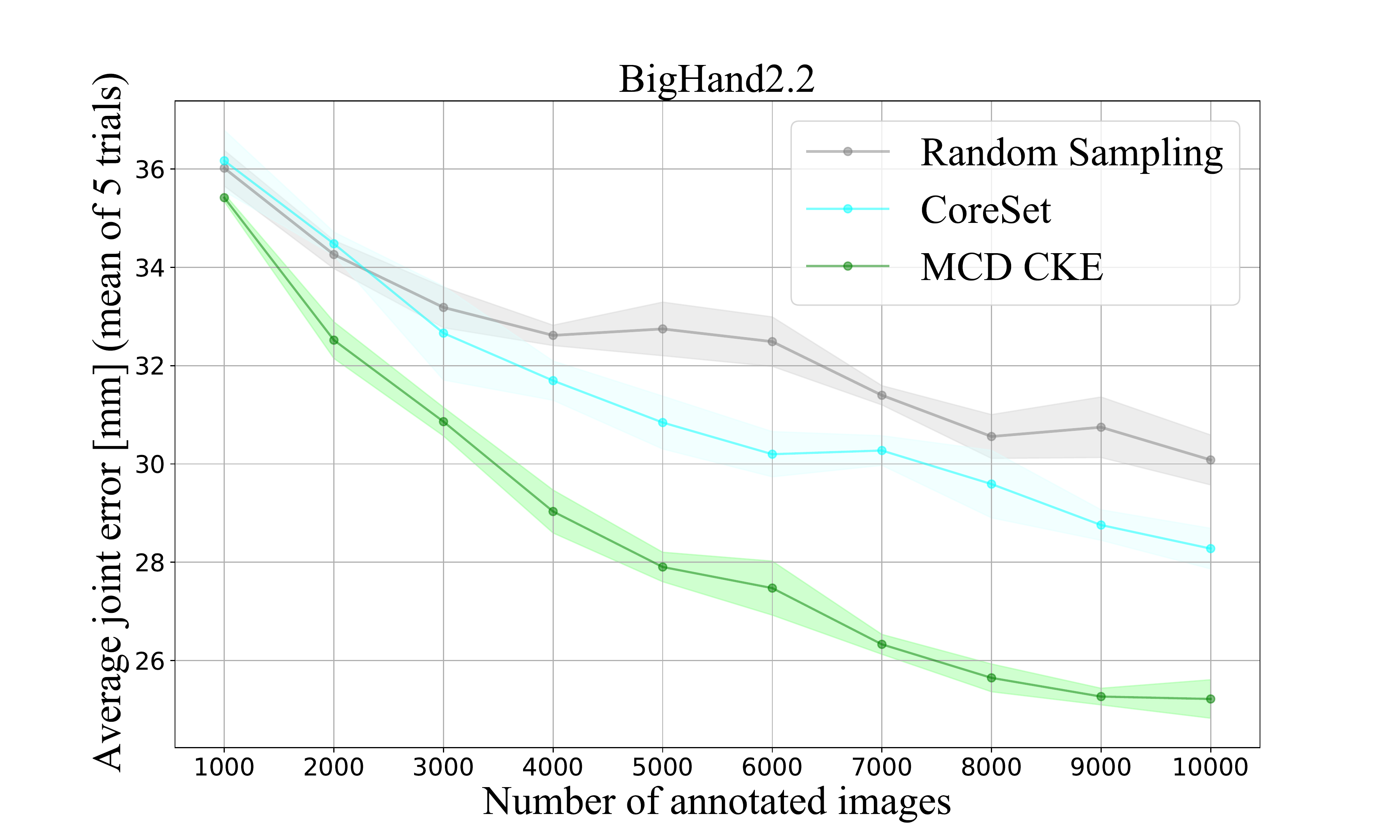}
    \includegraphics[trim=1cm 0.5cm 3.5cm 2.5cm, clip, width=0.33\textwidth]{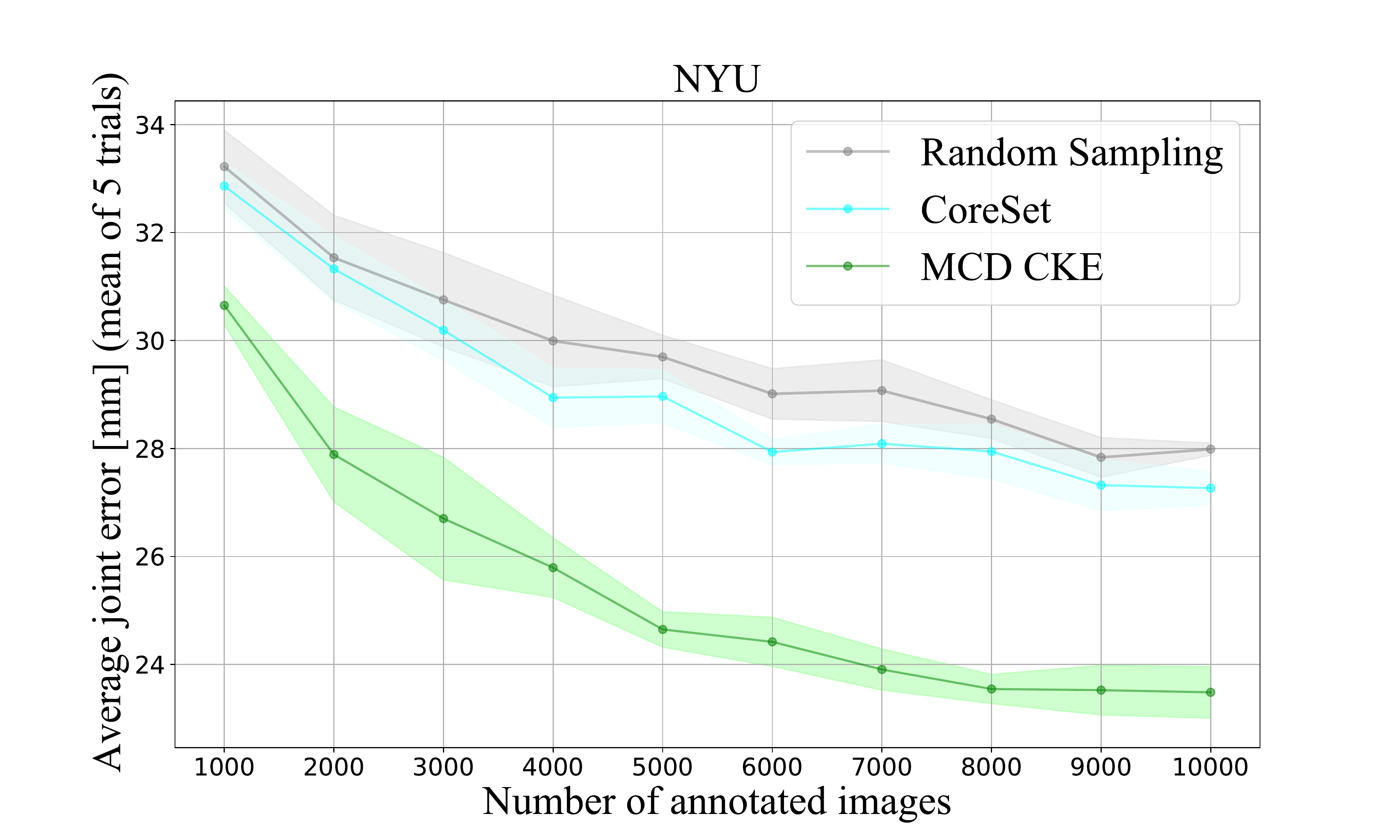}
    \caption{Empirical comparison I: Quantitative analysis of the proposed CKE method with Bayesian DeepPrior against the other methods applied on the standard version. Evaluated datasets: ICVL (Left), BigHand2.2 (Middle) and NYU (Right).}
    \label{fig:al_methods}
\end{figure*}
\begin{figure*}
    \centering
    \includegraphics[trim=1cm 0.5cm 3.5cm 2.5cm, clip, width=0.33\textwidth]{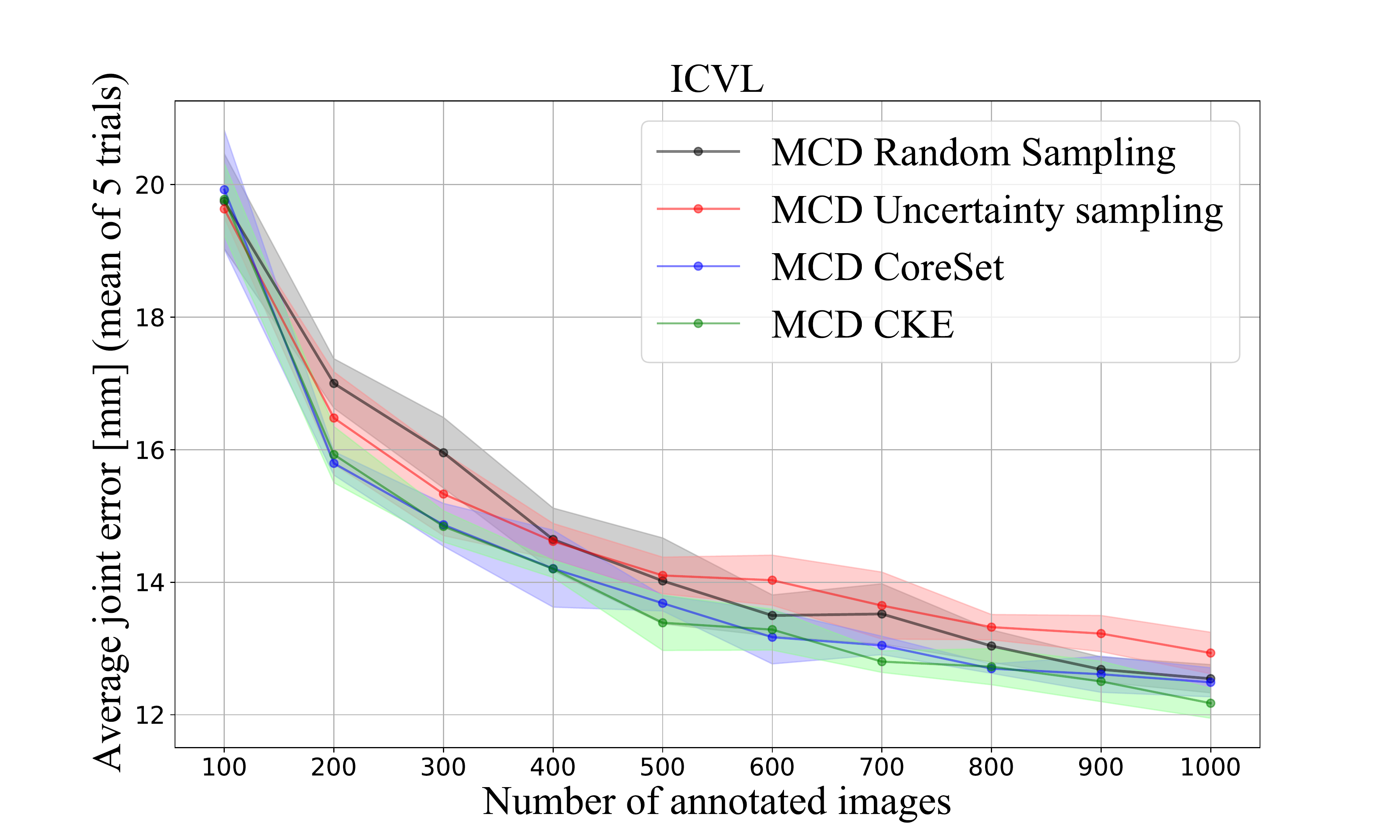}
    \includegraphics[trim=1cm 0.5cm 3.5cm 2.5cm, clip, width=0.33\textwidth]{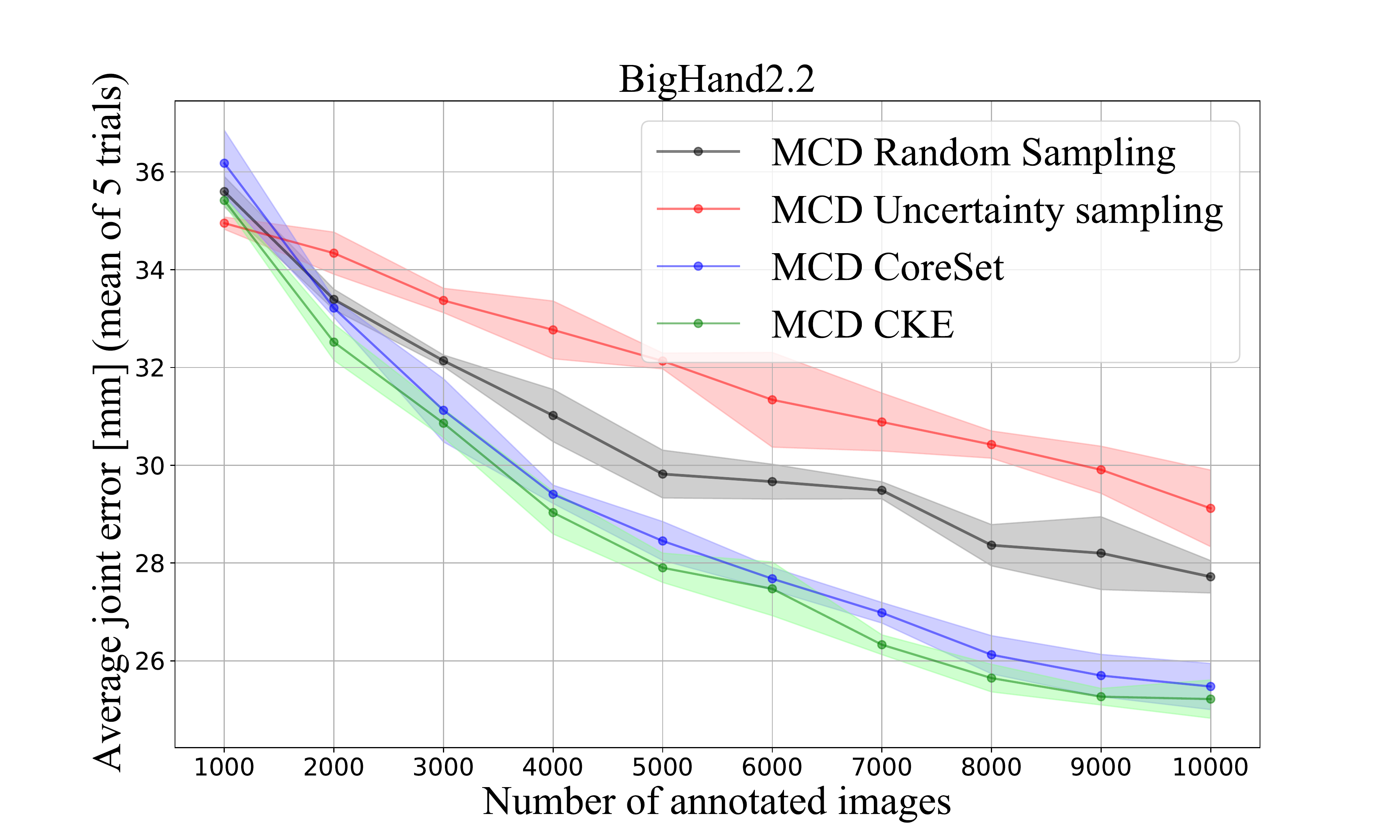}
    \includegraphics[trim=1cm 0.5cm 3.5cm 2.5cm, clip, width=0.33\textwidth]{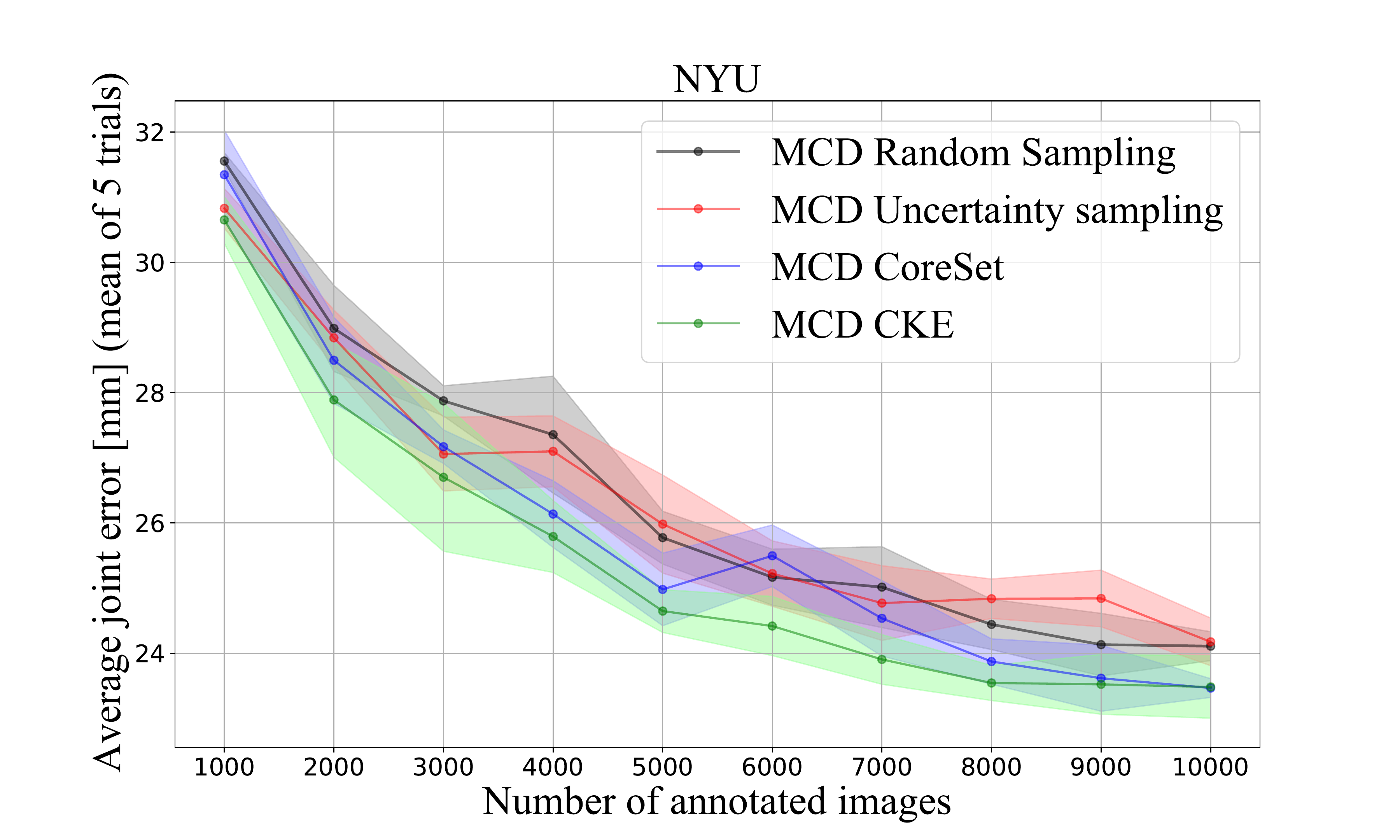}
    \caption{Empirical comparison II: AL performance comparison of the proposed CKE method against other MCD adapted samplers on ICVL(Left), BigHand2.2(Middle) and NYU(Right) data set.}
    \label{fig:mc_al_metthods}
\end{figure*}

As deducted in Kendall et al.~\cite{uncertainties}, we consistently obtain higher aleatoric variances than the epistemic ones. Hence, the noise present in the hand data tends to overcome the model's learning capability. Apart from BigHand2.2., the testing epistemic variance seems lower than on the training set. Furthermore, we notice an increased aleatoric variance on BigHand2.2 compared to the other sets. High noise during annotation could be the reason for it.
\subsection{Active Learning Evaluation}
% AL helpful in acquiring meaningful ex for the HPE : In an data collection process It reduces annotation time and provides, reduce the noisy annotated samples through aleatoric unc measurements.
Active learning has shown to be an effective tool \cite{Sener2017ActiveApproach,Sinha2019VariationalLearning,Sener2017ActiveApproach} in acquiring representative data for a learning model. Given the proposed Bayesian DeepPrior architecture, we presented in Section \ref{sec:cke} the CKE query method. While our pipeline gets an advantage from reducing the noisy samples through learnt aleatoric variances, we gather both geometric and epistemic information regarding new unlabelled poses. We shortly describe the active learning selection baseline used for both Bayesian and standard DeepPrior architecture:\\
\noindent \textbf{Random sampling}: The typical approach of annotating data just by uniformly sampling the unlabelled pool $U_{pool}$. \\
\noindent \textbf{Uncertainty sampling}: Although for 3D HPE regression there is no confidence measurement like in classification tasks, we apply this method only on the evaluated epistemic uncertainties. The unlabelled examples are inferred through the Bayesian DeepPrior and their epistemic variances result after 40 MCDs. For each predicted skeleton, we sum up all their corresponding epistemic deviations and we select to annotate the topmost uncertain hand poses.\\
\noindent \textbf{CoreSet}: It is one of the current state-of-the-art geometric acquisition function. Also, it has been widely used due to its task-agnostic properties relying either on the model's extracted features or on the output space. In our evaluation, we apply CoreSet directly on the predicted skeletons of both labelled and unlabelled sets. \\ 
\noindent \textbf{CKE}: This is our proposed method that extends the CoreSet functionality by including the epistemic deviations in the risk minimisation between the loss of the newly selected samples and the loss of the available labelled set.
% \end{itemize}
\begin{figure*}
    \centering
    \includegraphics[trim=0.1cm 0cm 0.0cm 0.0cm, clip, width=0.6\linewidth]{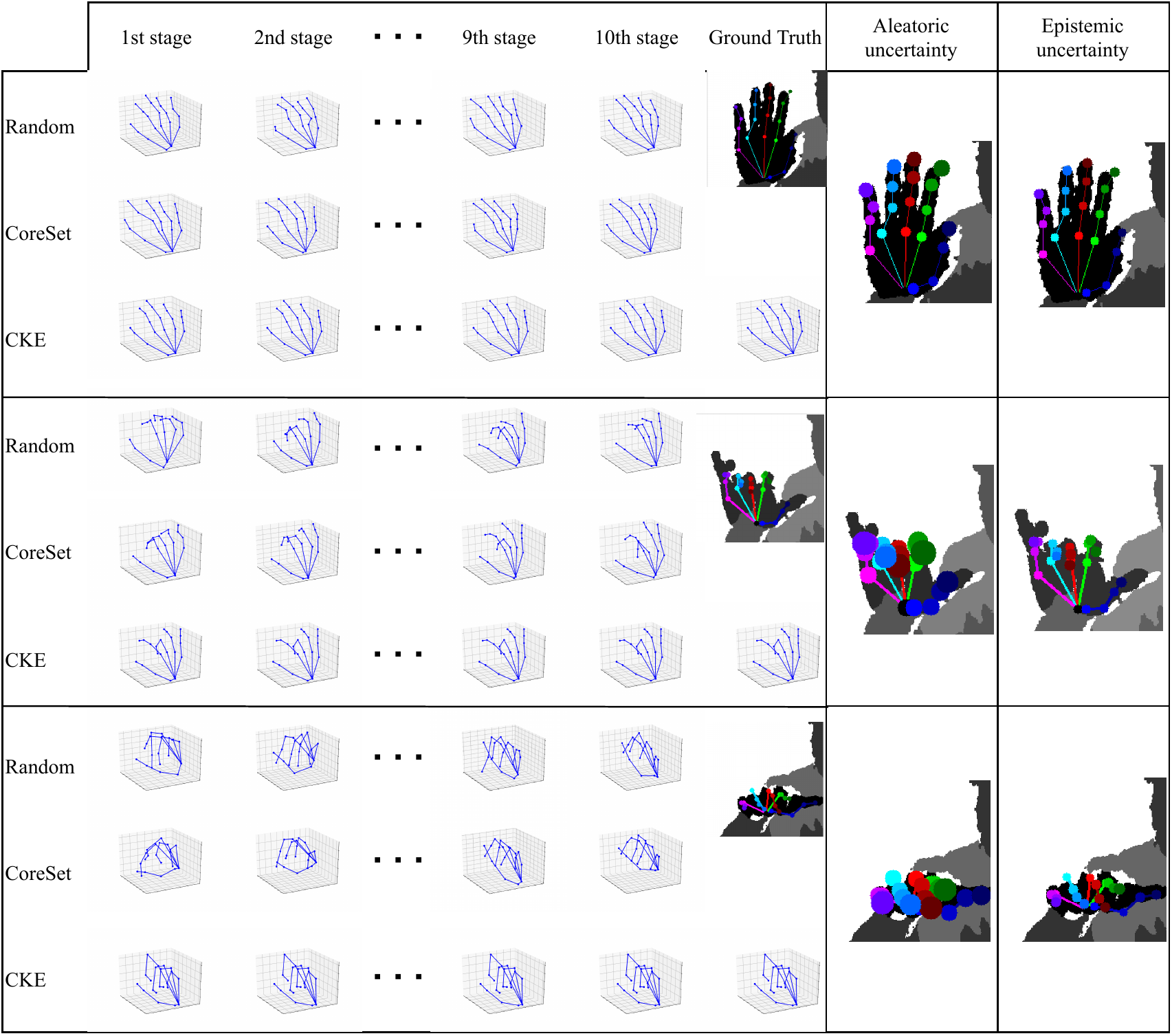}
    \caption{Qualitative comparisons on the Active Learning methods at different selection stages (Left). The 3D joint aleatoric and epistemic uncertainty variance (Right).}
    \label{fig:qual}
\end{figure*}

\subsubsection{Quantitative Results of Selecting Methods}
% Settings for the AL , U pool budget, how we split labelled/ unlabelled
In this part, we evaluate on three hand pose datasets: ICVL, NYU and BigHand2.2, under the pool-based scenario. We perform a quantitative comparison between the random sampling and CoreSet on the standard DeepPrior learner followed by CKE, our proposed selection method, on the Bayesian 3D HPE.
We follow the standard pool-based protocol, where there is a large pool, $U_{pool}$  of unlabelled data.  A small size initial set seed annotations $\s^0$  are made available for the first offline training of the target model. After this, we follow the same practice as in \cite{BeluchBcai2018TheClassification, Yoo2019LearningLearning} and randomly create a smaller pool $\s \subset U_{pool}$. Thus, we efficiently deploy the selection methods on $\s$ under the budget $B$. We repeat this systematically over 10 stages. As a quantitative metric, we quantify the performance in averaged MSE. Also, due to the variation in the initial selected set $\s^0$ and new random subsets $\s$, we average our results over 5 trials and while also computing the standard deviation.
% discuss Relate the performance over the entire training set

For NYU and BigHand2.2, we set a budget ($B$) as well as seed annotations $\s^0$ equal to 
$1000$ samples. Whereas, for ICVL, we set it $100$ due to its smaller size. 
% Also, we choose these dimensions when selecting the initial training set $\s^0$.
The intermediate subset $\s$ size used for active learning selection is set to 10\% of the entire $U_{pool}$  which results in 20,000 of BigHand2.2; 7,276 of NYU; and 1,601 of ICVL. In our CKE query function, we identified  the most informative samples when the uncertainty influence 
parameter was set to $\eta=0.3$.

% Figure \ref{fig:al_methods} illustrates a performance comparison between the proposed CKE selection method, random sampling and CoreSet which is also one of the state-of-the-art methods to date. As in the classic approach, we apply the other samplers (random sampling and CoreSet) to the standard version of DeepPrior in order to highlight its Bayesian approach.  \textcolor{red}{ We applied these selection techni to the standard DeepPrior sampler for highlighting the importance of the Bayesian adaptation. 
Figure~\ref{fig:al_methods} compares the performance of the proposed method with the default baseline (random sampling) and one of the state-of-the-art methods (CoreSet) in three different challenging data sets. Referring to the same Figure, we can clearly see that the proposed method consistently outperforms the existing state-of-the-art. Specifically, after 10 AL passes we achieve  the lowest averaged MSE on every benchmark accordingly: 12.17 mm for ICVL, 23.48 mm for NYU and 25.21 mm for BigHand2.2. This demonstrates how effective the Bayesian AL is in obtaining top accuracy with fractions of the entire datasets.

% Quantitatively, we show a consistent gap between CKE and CoreSet throughout the selection stages for all hand datasets. Thus, we achieve after 10 AL passes the lowest averaged MSE on  every benchmark accordingly: 12.17 mm for ICVL, 23.48 mm for NYU and 25.21 mm for BigHand2.2. When comparing to the performance obtained using the entire training set, our CKE method managed to maintain a closer 3D hand joint error while using small fractions of data. In particular, for ICVL we gained 2 mm errors with only 1000 samples, while with 5\% informative data of BigHand2.2, the accuracy was affected just by 3 mm. The most significant result was yielded at less than 14\% of the NYU training set where we noted a difference of 1 mm.

% we gained just 2 mm testing set errors for ICVL with only 1000 samples. We also observe this behaviour when using under 5\% of the BigHand2.2 and a 3 mm error increase. The best achievement resulted for NYU where 1 mm is lost at less than 14\% from the original training set.  

% This demonstrates that with reduced meaningful 3D hand poses we can obtain desired performance when acquiring and annotating a dataset.

% Quantitative comparison bet

Similarly, Figure~\ref{fig:mc_al_metthods} illustrates the performance of the  existing selection methods and our proposed sampling technique CKE when applied with the Bayesian DeepPrior. Comparing the performance of Coreset when the learner is DeepPrior (Figure~\ref{fig:al_methods}) vs when the learner is Bayesian DeepPrior (Figure~\ref{fig:mc_al_metthods}), we observe the improvement of the averaged MSE  on every dataset. This trend is followed even in random sampling technique. Hence, this demonstrates that modeling aleatoric and epistemic uncertainties while training learner in the AL framework is effective.  
% the proposed method when the learner is the Bayesian DeepPrior. We can observe that  the Bayesian DeepPrior, the proposed learner converges faster with lower joint error than the standard DeepPrior(Compare Figure~\ref{fig:al_methods}. We generated two separate plots for sanity.) in every selection methods. However, the combination of the  Bayesian DeepPrior
% as learner and the CKE as the sampler attains the best performance in all the three benchmarks. 
% This demonstrates the effectiveness of the proposed methods over the counter-part and existing methods.
% The Uncertainty sampling selection seems to perform poorer even than random. This effect can occur if too many hard samples are used since the beginning of the query process. Thus, the model will not be able to discriminate simple frontal hand poses or with a reduced degree in articulations. 
% Nevertheless, our CKE function maintains the best performance while showing the benefits in combining CoreSet geometric approach with the MCD epistemic uncertainty. These insights indicate a great potential in deploying a hand pose-representative acquisition system.

% \begin{figure*}
%     \centering
%     \includegraphics[width=0.49\textwidth]{figures/nyu.pdf}
%     \includegraphics[width=0.49\textwidth]{figures/nyu_mc.pdf}
%     \caption{Caption}
%     \label{fig:my_label}
% \end{figure*}
\subsubsection{Qualitative comparison}
We also evaluate the sampling methods qualitatively. We compare the baselines (random and CoreSet) against our proposed CKE function. To this end, we extract the predicted 3D hand skeletons on NYU test set and track their structural representation in the first two initial stages as well as in the last two AL stages. In the left part of Figure \ref{fig:qual}, we can observe that CKE under the Bayesian DeepPrior learner generates 3D hand poses quite closer to the ground truth from the early stages. Whereas, the other two methods failed to do so. These characteristics are equally visible even in highly articulated poses as shown in the last row.

From a qualitative perspective, we also evaluate the relevance of the epistemic and aleatoric uncertainty values in the context of 3D joint error locations. 
% As introduced in Section \ref{bayesian}, the Bayesian DeepPrior outputs both types of uncertainty for each 3D joint coordinate. Thus, for simplicity, we define the the joint radius by it maximum aleatoric or epistemic deviation. This definition is visually shown in Figure \ref{fig:qual}(Right).
% After computing the uncertainties mean values in Section \ref{ep_al}, we can observe how mainly the aleatoric uncertainty changes with different hand pose. 
The last two columns of Figure~\ref{fig:qual} depict these uncertainties. 
The extended variance present at the fingertips (proportional to the radius of the circle) can be interpreted as high acquisition noise. Moreover, we can observe that for occluded poses (middle and last row), the bigger circle indicates that the model is less confident in such cases. This happens
when there are not sufficient training examples in such extreme poses and occlusions.
% due to the excessive non-linearity when mapping with the reviewed 3D-HPE. 
To overcome this, a more powerful estimator trained on annotated examples with such extreme poses and occlusion needs to be deployed.
\section{Conclusions}
We have elaborated the first work of AL applied to the 3D hand pose estimation task. We successfully approximated the 3D-HPE DeepPrior as a BNN while deriving its model and data-dependent uncertainties. We have shown through qualitative and quantitative evaluation how the two uncertainties play a critical role in the AL scheme. Furthermore, by combining the geometric sampling from CoreSet, we proposed a sampling technique, CKE, suitable for Bayesian DeepPrior infrastructure. Under the pool-based scenario, we achieve the lowest 3D joint errors with the least amount of data for three well-know datasets. To conclude, this work demonstrates the lack of representativeness and redundancy that can be present when gathering a 3D hand dataset. Therefore, a Bayesian approximation together with the CKE acquisition method may help in building a holistic and model-refined dataset while saving a considerable annotation time. 
\label{conclusions}
\section*{Acknowledgement}
\vspace{-0.2cm}
This work is partially supported by Huawei Technologies Co. and by EPSRC Programme Grant FACER2VM (EP/N007743/1). We also like to thank Anil Armagan for his insights and discussions.
% \clearpage
{\small
\bibliographystyle{ieee_fullname}
\bibliography{al}
}

\end{document}